\documentclass[10pt,twocolumn,letterpaper]{article}

\usepackage[pagenumbers]{cvpr}

\usepackage{algorithm}
\usepackage{algpseudocode}
\usepackage[dvipsnames]{xcolor}

\definecolor{cvprblue}{rgb}{0.21,0.49,0.74}
\usepackage[pagebackref,breaklinks,colorlinks,citecolor=cvprblue]{hyperref}
\usepackage{tabularx}
\usepackage[capitalize]{cleveref}
\crefname{section}{Sec.\ }{Secs.\ }
\Crefname{section}{Section}{Sections}
\Crefname{table}{Table}{Tables}
\crefname{table}{Tab.}{Tabs.}

\title{SynCellFactory: Generative Data Augmentation for Cell Tracking}

\author{
Moritz Sturm\thanks{Interdisciplinary Center for Scientific Computing (IWR), Heidelberg University.} \\
{\tt\small mosturm@gmx.de}
\and
Lorenzo Cerrone \footnotemark[1] \\
{\tt\small lorenzocerrone@gmail.com}
\and
Fred A. Hamprecht \footnotemark[1] \\
{\tt\small fred.hamprecht@iwr.uni-heidelberg.de}
}

\begin{document}
\maketitle

\begin{abstract}
\noindent Cell tracking remains a pivotal yet challenging task in biomedical research. The full potential of deep learning for this purpose is often untapped due to the limited availability of comprehensive and varied training data sets. In this paper, we present SynCellFactory, a generative cell video augmentation.

\noindent At the heart of SynCellFactory lies the ControlNet architecture, which has been fine-tuned to synthesize cell imagery with photorealistic accuracy in style and motion patterns. This technique enables the creation of synthetic yet realistic cell videos that mirror the complexity of authentic microscopy time-lapses.

\noindent Our experiments demonstrate that SynCellFactory boosts the performance of well-established deep learning models for cell tracking, particularly when original training data is sparse.
\end{abstract}

%% --- Introductuon ---
\section{Introduction}
\label{sec:intro}
Digital time-lapse microscopy allows for large-scale observations of cells over time, providing a deeper understanding of cellular processes \citep{TLM,TLM2,TLM3}.
However, to fully harness the potential of time-lapse imaging,  automated cell tracking approaches are needed, which can provide a quantitative analysis of cell behavior for vast amounts of data. 

Cell tracking is characterized by challenges such as variable image contrast, intricate behaviors such as cell division and, in some examples, indistinguishability of cells. Recent advancements in computer vision have shown that neural networks are highly effective in multi-object tracking tasks \citep{MOT1,Trackformer,MOT2}. However, their application in cell tracking remains limited and exploratory, primarily due to the scarcity of annotated cell tracking data. This shortage hinders the development of robust neural network models for cell tracking \citep{CTC_23}.

In recent years, although medium-scale annotated tracking data sets in the order of several thousand timeframes have become available \citep{Caliban,DeepSEA,CTC_23}, they are limited to specific cell styles. However, given the vast diversity of cell types and imaging techniques, it is difficult for a single data set to cover all these variations comprehensively.
The reason for the lack of a broad range of annotated training cell tracking data is the costly expert annotation combined with cell tracking's niche status in the broader field of machine learning.

To address these challenges, this paper introduces \textit{SynCellFactory}, a generative data augmentation strategy specifically designed for cell tracking. 
We embrace the power of conditioned 2D diffusion models \citep{SD_orig,CNet} to generate high-quality synthetic cell videos that mimic the appearance and behavior of real cell data sets. 
% But more importantly, to our knowledge, this is the first solution that can be applied as generative data augmentation without any prior biological knowledge.

Utilizing as little as one annotated cell video for training, \textit{SynCellFactory} can generate an extensive library of annotated videos in a consistent style, effectively augmenting the available training data. This advancement holds the potential to transform cell tracking by enabling the application of sophisticated deep learning models previously hindered by data scarcity.
Our research focuses on evaluating the impact of this data augmentation on the performance of state-of-the-art cell tracking models. The results indicate that this novel approach addresses the data scarcity issue and enhances tracking accuracy. We aim to open the door for large-scale deep learning architectures in cell tracking. 
\\

\noindent To summarize, our contributions are: 
\begin{itemize}
    \item \textit{SynCellFactory}, a generative data augmentation pipeline for cell tracking. 
    \item The proposed pipeline demonstrates out-of-the-box robust results on a wide variety of data sets without complex hyperparameter tuning or domain-specific knowledge. 
    \item Empirical proof that the proposed data augmentation strategy can further enhance accuracy of an already leading deep learning cell tracking method. 
\end{itemize}

\begin{figure*}
  \centering
      \includegraphics[width=1.\linewidth]{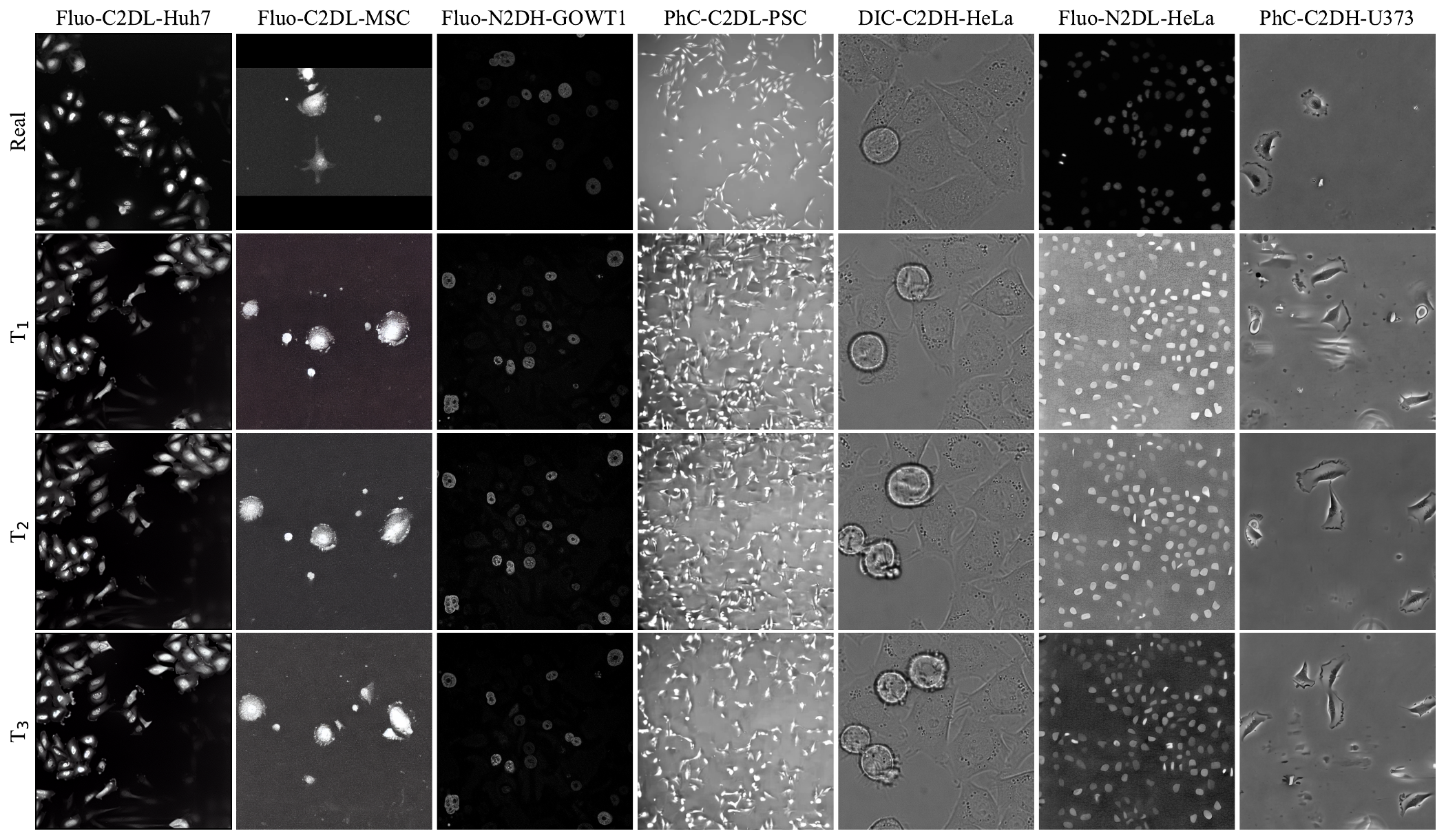}
  \caption{Showcase of real (top row) and synthetic (other rows) images generated using \textit{SynCellFactory}. The training data sets are a subset of the  2D Cell Tracking Challenge \cite{CTC_old, CTC_23} and provide a broad spectrum of cell lines and microscopy modalities.}
  \label{fig:galleria}
\end{figure*}

\begin{figure*}
  \centering
      \includegraphics[width=0.95\linewidth]{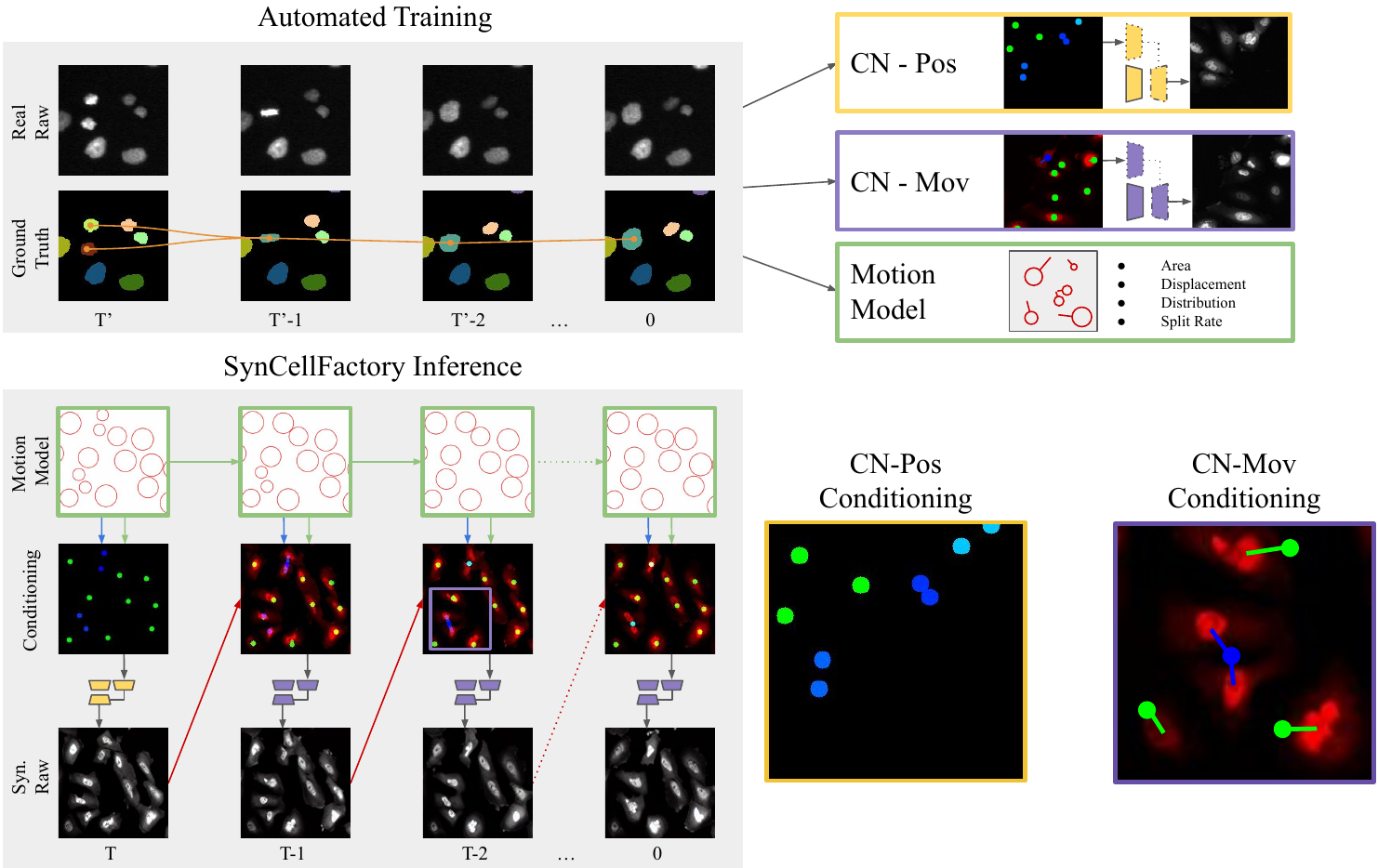}
  \caption{\textit{SynCellFactory} is a data augmentation pipeline designed to create unlimited high-quality synthetic raw video data and corresponding pseudo ground truth. It trains three key components using a small, and possibly sparsely labeled data set: Positional ControlNet (CN-Pos), Movement ControlNet (CN-Mov), and a 2D movement engine for realistic simulations.
The process initiates in reverse, with the motion model generating a conditioning image at time $T$ for CN-Pos. This image illustrates the expected centers of cells using colored dots, where each color signifies a specific cell state in the mitotic cycle: green during the interphase and blue during the different phases of  cell division.
CN-Pos then employs this information to generate a realistic frame for time $T$. Subsequently, CN-Mov assumes the role of producing the next frame $T-1$, using as conditioning an RGB image that combines the previously generated frame (in the red channel) with the projected positions and movement patterns (in green and blue channels). Derived from the motion model, these patterns represent each cell's trajectory from its current to its anticipated next position as a line connecting the two.
By iteratively applying CN-Mov, \textit{SynCellFactory} can efficiently produce time-lapse sequences of any desired length, suitable for training deep learning pipelines in cell tracking.}
  \label{fig:main_scf}
\end{figure*}

\subsection{Related Work}
\label{sec:formatting}
\textbf{Generative Data Augmentation:} Data augmentation is ubiquitous in deep learning and is known to help enhance the robustness and generalization of deep learning models \cite{improve_robust,img_data_aug_dl,Rescon,bprcnn}. 
Most data augmentation strategies in computer vision are based on simple transformations of the input data, including rotations, non-rigid deformations, color alteration, and cropping of images or videos. Recently, approaches based on generative machine learning are gaining traction \citep{syn_diff}. The idea is to use a generative model to sample a new data point drawn from the estimated real data distribution. 
Until recently, Generative Adversarial Networks (GANs) were the primary choice for this task \citep{GAN,pop_gans1,pop_gans2}. However, diffusion models advanced and eventually surpassed GANs in image synthesis quality \citep{Diff_bt_GAN}, making diffusion models the new state of the art for generating synthetic image data \citep{syn_diff, robust,he2023}.
Refs.~\citep{syn_diff,he2023,eff_diff,Mert} demonstrate the effectiveness of generative data augmentation using diffusion models in terms of achievable image classification accuracy. 
These advances also start to impact the medical imaging domain, where they show great promise  \citep{nice_med_syn1,nice_med_syn2,nice_med_syn3,monai} and have already proven effective when used to train deep learning models \citep{Syn_Data_in_med, enhancing1}. The main motivation behind using generative data augmentation in medical computer vision is the scarcity of ground truth as well as privacy concerns.
%Motivated by these advances and the real need for larger data sets in the medical domain, we developed our pipeline to combat data shortage in annotated data sets for cell tracking. 

\textbf{Cell Synthesis:} Several approaches to synthesizing artificial cell images have been proposed since the late 90s~\citep{late90}. 
Early methods relied heavily on handcrafted parametric models. These models, informed by extensive biological knowledge, simulated the appearance and temporal evolution of cells~\citep{MitoGen}.

However, with the evolution of deep learning, an effort has been made to reduce the domain-specific knowledge required to generate realistic cell video.
While \cite{cell_cycle} combines statistical shape and evolution models with conditional GANs to generate 2D+t data, \cite{cell_video_diff} employs video diffusion models to create full cell videos directly from the training set.
Most existing models focus on generating a single cell in either 2D or 3D \citep{Cell_syn_20,wavelets,3D_GANs,GAN,isolate_cell}. To simulate a cell population, some methods first model single cell appearance and temporal development and then incorporate simultaneous evolution of multiple cells to model a population of cells \citep{MitoGen,cell_cycle}. Our proposed method is based on a similar premise to model cell appearance and temporal development separately, but distinguishes itself by modeling the image of an entire cell population jointly.
%Both \cite{cell_cycle} and \cite{MitoGen} adopt a three-phase process for cell generation: creating a cell phantom with a statistical shape model, synthesizing internal texture, and simulating the imaging system with point spread function and Gaussian noise. In contrast, our method bypasses these steps using ControlNet \cite{CNet}, simultaneously generating cell shape, texture, and sensor noise. 
Closely related to our work, \cite{cell_video_diff} generates an entire video at once using a 3D diffusion model guided by optical flow; crucially, their approach does not produce pseudo ground truth labels, while \textit{SynCellFactory} does. Ref.~\citep{as_ist} proposes a CycleGAN \cite{cyclegan} based model which can also generate raw data as well as lineage and segmentation pesudo ground truth. An important difference is that said work relies on simulated ground truth segmentation masks to condition the data generation; this is a much harder problem to solve than the simple trajectory modeling employed here. 

%% --- Method ---
\section{Method}
\textit{SynCellFactory} operates on the principle of decoupling cell dynamics from their appearance. By controlling these two aspects independently, our model can generate highly heterogeneous annotated cell videos.
Our model comprises two principal components:
\begin{enumerate}
\item A simple 2D motion model that simulates the spatial distribution and dynamics of cell cultures. It uses statistical parameters derived from real data, enabling \textit{SynCellFactory} to produce coherent and physically plausible timelapses, irrespective of frame count. This simulation approach enhances the diversity and realism of the generated data.
\item Two distinct ControlNets, each with a specific function, are trained for photorealistic rendering. The first, CN-Pos, is adept at inpainting cells with lifelike appearance at accurate spatial positions. The second, CN-Mov, focuses on the temporal displacement of individual cells across consecutive frames, ensuring temporal consistency. Particularly crucial for our use case in data augmentation, these ControlNets demonstrate efficient training capabilities, even with limited data such as a single time-lapse sequence, making them well suited for augmenting data sets in cell tracking applications.
\end{enumerate}

In the following sections, we will detail the methodology of our model and introduce an automated protocol for training \textit{SynCellFactory} on new data sets.

\subsection{Motion Model}
\label{sec:phys}
Cell movement under a microscope is a process influenced by various factors like cell type, environmental conditions, and cell-to-cell interactions. Biologically accurate modeling of the cell behavior is outside the scope of this manuscript. Therefore, we are not implementing intra-cellular forces and complex behaviors such as amoeboid, ciliary, and flagellar motions. The goal of our motion model is to generate plausible spatial cell configuration and displacement, leaving the heavy lifting of visually representing cell interactions to the generative neural network. 

Our engine represents cells as 2D disks, capable of moving, splitting, and interacting with one another.
The motion model initializes a population of cells with randomly sampled positions and sizes, drawn from the distribution observed in real cells from our training data. This population dynamically evolves following a stochastic Brownian motion, guided by statistics extracted from annotated real cell videos. These key parameters include the mean and standard deviation of cell area ($A_c$ and $\sigma_A$) and the cell displacement distribution, which we model using a gamma distribution $G(\alpha_d,\Theta_d)$. These statistical models are derived from available segmentation and tracking ground truth data.
A significant advantage of our approach is its flexibility in adjusting the tracking difficulty of the generated data sets. By manipulating variables such as the number of cells, their movement speed, and the frequency of cell splitting events, we can tailor the complexity of the tracking task. This allows us to create data sets that either exceed or are comparable in complexity to the original data, accommodating a range of research and application needs.
Details on the statistical methods, the specific equations governing cell movement, and the pseudocode for the simulation are provided in supplementary Sec.~A. 

\subsection{ControlNet}
ControlNet \cite{CNet} is a popular architecture for enabling the conditioning of text-to-image generative models like stable diffusion \cite{git_SD}.

Our ControlNet uses the standard pre-trained Stable Diffusion v1.5 backbone trained on natural images. In contrast to the original ControlNet model that maintains the stable diffusion decoder locked, we fine-tune it to the appearance of biological images. 

Morever, training a ControlNet in the context of latent text-to-image diffusion models involves triplets of the form $(c_{\text{txt}},c_{\text{img}}, i_{\text{tgt}})$, where $c_{\text{txt}}$ is a conditioning text, $c_{\text{img}}$ is the  conditioning image and $i_{\text{tgt}}$ is the target image.
The text conditioning has been kept fixed for all our experiments to the prompt ``cell, microscopy, image'' to provide a good initialization during the early stages of training. 

In the following sections, we describe the specific details and  $c_{\text{img}}$ conditioning the CN-Pos and CN-Move models. 

\begin{figure}
\centering
    \includegraphics[width=0.95\linewidth]{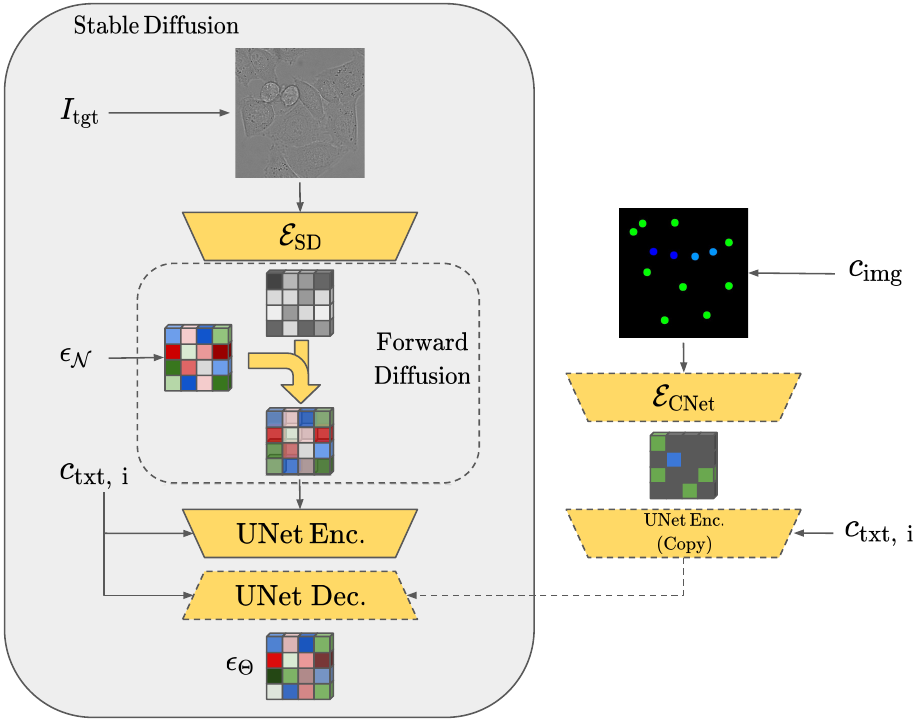}
\caption{Train-time computation flow in the ControlNet \cite{CNet}. Initially, the source image $I_\text{tgt}$, and the conditioning input $c_{\text{img}}$ are transformed into $4\times64\times64$ embeddings through two convolutional encoders: $\mathcal{E}_{\text{SD}}$ for the source and $\mathcal{E}_{\text{CNet}}$ for conditioning. In the stable diffusion training routine, the source embedding undergoes a forward diffusion process, where Gaussian noise is incrementally added. Subsequently, a UNet encoder-decoder attempts to estimate and revert this noise perturbation. This is accomplished by applying the mean square error loss between the input noise $\epsilon_{\mathcal{N}}$ and the decoder output $\epsilon_{\Theta}$. Unique to ControlNet architecture, additional image conditioning is integrated via an auxiliary branch. In the diagram, neural network blocks with solid lines represent components with fixed parameters during training, while those with dotted lines indicate blocks subject to finetuning. The described architecture is used in both CN-Pos and CN-Mov.}
\label{fig:cnet_detailed}
\end{figure}

\subsubsection{Positional ControlNet}
\label{subsec:cn-pos}
The positional ControlNet CN-Pos is tasked with drawing realistic looking cells conditioned on a given position (see \cref{fig:main_scf}). 
This model plays two fundamental roles in \textit{SynCellFactory}: it is used to generate the last frame of our synthetic videos and is the pre-trained backbone of the CN-Move model.
At train time, the position map is constructed by computing the center coordinates for each cell given the corresponding detection ground truth (see \cref{fig:main_scf}). Each detection is represented in the conditioning image $c_{\text{img}}$ as a disk with a fixed radius of $r=\sqrt{{A_c}/{\pi}}/4$, where $A_c$ is the data set average cell area in pixels. 
Fixing the radii of the disks instead of varying them proportionally to the cell size empirically improved the diversity of cell appearances.
Moreover, the disks are colored according to the stage in the cell cycle. The disk is green while the cell is in the interphase, linearly changing between green and blue during the Mitosis phases (Prophase, Metaphase, Anaphase, and Telophase) and linearly reverting from blue to green after the cytokinesis phase and the split event is complete. The cycle length is a hyperparameter that needs to be set according to the specific cell type for each data set.
At inference time, the conditioning image $c_{\text{img}}$ is derived by the state simulated by the motion model, where the center of the simulated cells are converted into color-coded disks. 

\subsubsection{Movement ControlNet}
CN-Mov is analogous to CN-Pos in its implementation details; the sole difference lies in its conditioning. 

CN-Mov is tasked with predicting frame $t-1$ conditioned on the frame at time $t$, the position map at time $t-1$, and the displacement vectors.
Frame $t$ is encoded in the red channel of $c_{\text{img}}$, while the position map at time $t-1$ is encoded in the blue and green channels as described in \cref{subsec:cn-pos}. 
Displacement vectors describe the movement of the cells between frames and are encoded as lines connecting the center position of cells. A more detailed description of the conditionings can be found in supplementary Sec.~B.

\subsection{SynCellFactory Inference}
During the sampling process, we apply the trained ControlNets on the output of our motion module to generate realistic-looking videos (see Fig.~\ref{fig:main_scf}).  
The choice to sample our video in reverse time is arbitrary, but simplifies network learning: in backward time, cell merge events require the network to predict only a single position from two merging cells. Conversely, in forward time, the network would face the more complex task of generating two new cells from one during a split event.

The inference process is initiated with CN-Pos generating frame $t$. Subsequently, CN-Mov iteratively takes the generated frame $t$ and, in conjunction with the motion model, samples a new frame $t-1$. This iterative process continues until the desired video length is reached.
In all our experiments, we generated a fixed length of 12 frames.

\subsection{Segmentation Pseudo Ground Truth}
\label{subsec:seggt}
\textit{SynCellFactory} produces only raw video, detection, and lineage ground truth, but critically is not able to produce instance segmentation ground truth. This is a significant challenge, as several deep learning methods \citep{kit_track,DeepSEA} require full segmentation for effective training. To address this, we rely on Cellpose \cite{cp1,cp2}, a renowned deep learning framework for cell segmentation, to create pseudo-ground truth segmentation and thereby mitigate this limitation.
Cellpose is particularly effective due to its range of pretrained models that cater to various cell image styles, boasting high segmentation accuracy right out of the box. For our cell data sets, we chose the pretrained model recommended by Cellpose, and further finetuned it for 100 epochs using ground truth segmentation masks from our training data. This finetuning allows the model to more closely adapt to the unique characteristics of a particular data set.
While the segmentation labels produced by Cellpose are mostly accurate, they occasionally miss or hallucinate cells. To improve upon this, we integrated the motion model into the segmentation process. In cases where a segmentation mask does not overlap with any part of the generated detection ground truth, it is removed. Conversely, when a ground truth detection is present without a corresponding segmentation mask, we generate one by drawing a circle with a radius $r=\sqrt{{A_c}/{\pi}}$. This step is crucial to ensure that small inaccuracies or hallucination artifacts produced by the ControlNet do not receive segmentation masks, while cells that are challenging to segment are still provided with an approximate mask. 
See supplementary Sec.~C for an illustrative example of the correction step.
%An illustrative example of this segmentation correction process can be seen in Fig. \ref{fig:corr}.

\subsection{Automated Training}
To reduce the domain-specific expertise required to train \textit{SynCellFactory}, we propose a fully automated pipeline for training the ControlNets and the sampling from the motion module. 

The only required hyperparameters that have to be manually specified to produce our synthetic videos are: the number of videos to be generated, the number of frames per video, and the characteristic length of the mitosis cycle. 

Other parameters,such as movement statistics for the motion model, are automatically inferred from the raw data and ground truth annotations. One of the main challenges with training the ControlNet is defining a good stopping criterion. We found that in our applications, the number of required steps to convergence is correlated with the average cell number in each frame. Thus, we can automatically set an appropriate number of training iterations for our ControlNets (see supplementary Sec.~D)

%% --- Experiments and Results ---
\section{Experiments and Results}
\subsection{Data sets}
To validate the proposed \textit{SynCellFactory} we use the publicly available Cell Tracking Challenge (CTC) \citep{CTC_old,CTC_23} \footnote{\url{http://celltrackingchallenge.net/2d-datasets/}} data sets.
The CTC is an open benchmark to test new cell tracking and segmentation advancements on data sets with a large variety of cell types, microscopy modalities, and different styles, as well as inhomogeneous spatial and temporal resolution.

For our experiments, we focus on the seven 2D data sets enumerated in \cref{fig:galleria}.
%\footnote{data set names indicate the microscopy meathod used: 'Fluo' %for Fluorescence microscopy, 'PhC' for Phase Contrast microscopy, and %'DIC' for Differential Interference Contrast microscopy.} 
Each data set consists of two timelapses with full tracking annotations of ground truth and only partial hand-curated segmentation. The tracking ground truth includes detection and identity masks for each frame and the corresponding cell lineages.
The number of frames per video ranges from 30 (Fluo-C2DL-Huh7) to 300 (PhC-C2DL-PSC).

Our experiments used a single time-lapse for the training and validation and one for testing tracking accuracy. 
Sample still frames from the mentioned data sets can be found in \cref{fig:galleria}.
We have elided from the pool of data sets for our experiments Fluo-N2DH-SIM+ because it is an already simulated cell video \citep{MitoGen}. We excluded BF-C2DL-HSC and BF-C2DL-MuSC because the boundary conditions of the petri dish could not be reproduced by our motion model and because these data sets already provide several thousand annotated frames, rendering augmentation less relevant.

\subsection{Experimental Setup}
The training of CN-Pos and CN-Move follows the standard procedure as presented in \cite{CNet}; the only major difference is that we also finetune the stable diffusion UNet decoder block. In the original ControlNet architecture, this module is fixed at train time. %Still, since our application domain is very different from the natural images of the original work, we found it beneficial to finetune this block. 
A detailed diagram of the training computation is shown in \cref{fig:cnet_detailed}.

Since we could only access a single annotated timelaps for training our ControlNet, we relied on data augmentation. In particular, we found random cropping of the images and random 90 degree rotations beneficial. The training process is highly resource-intensive, roughly 20 hours on a Nvidia A100 with 40GB of VRAM. A more detailed explanation of the ControlNet training protocol is reported in supplementary Sec.~D.

To evaluate the usefulness of \textit{SynCellFactory} as a data augmentation strategy, we used it in conjunction with the state-of-the-art deep learning model EmbedTrack \citep{kit_track}.
EmbedTrack uses CNNs to predict cell segmentation and tracking jointly. 
For training EmbedTrack, we use the default settings suggested by the authors; in particular, for each experiment, we trained the model for 15 epochs using the CyclingAdam optimizer \cite{ADAM}. For testing, we used the best-performing model with respect to the validation intersection over union.

\subsubsection{Tracking Metric}
To evaluate the tracking performance of our trained models, we use the official tracking accuracy measure (TRA) provided by the Cell Tracking Challenge \citep{ctc_metric}.

The TRA score is based on the concept of Acyclic Oriented Graph Matching AOGM. Suppose we represent the tracking predictions as an acyclic-oriented graph. In that case, we can measure the accuracy of a predicted graph by counting the number of operations required to transform the prediction into the ground truth graph.
The operations considered are node insertion, node deletion, edge insertion, and edge deletion. 
Formally, the TRA score is defined as:
\begin{equation}
\text{TRA} = 1 - \frac{\text{min}\left( \text{AOGM}, \text{AOGM}_0 \right)}{\text{AOGM}_0},
\end{equation}
where AOGM represents the weighted sum of operations required to build the prediction graph and $\text{AOGM}_0$ represents the weighted sum of operations required to build the ground truth graph. Higher TRA scores indicate better tracking performance.

\begin{table*}
\centering
\begin{tabular}{@{}l c c c@{}}
\toprule
data set         & W/o \textit{SynCellFactory} & With \textit{SynCellFactory} & $\alpha$ \\ 
\midrule
Fluo-C2DL-Huh7  & 0.960 $\pm$ 0.002   & \textbf{0.966} $\pm$ 0.003   & 0.66  \\
Fluo-C2DL-MSC   & 0.624 $\pm$ 0.060   & \textbf{0.685} $\pm$ 0.060   & 0.50   \\
DIC-C2DH-HeLa   & 0.968 $\pm$ 0.001   & \textbf{0.974} $\pm$ 0.001   & 0.80   \\
Fluo-N2DH-GOWT1 & 0.980 $\pm$ 0.003   & \textbf{0.989} $\pm$ 0.002   & 0.48  \\
Fluo-N2DL-HeLa & 0.939 $\pm$ 0.002 & \textbf{0.981} $\pm$ 0.002 & 0.87 \\
PhC-C2DL-PSC & 0.958 $\pm$ 0.001 & \textbf{0.960} $\pm$ 0.001 & 0.20 \\
PhC-C2DH-U373   &   \textbf{0.938} $\pm$ 0.008 &  0.935 $\pm$ 0.007  &  0.87   \\
\bottomrule
\end{tabular}
%}
\caption{Tracking Accuracy Measure TRA (higher is better) obtained with and without our \textit{SynCellFactory}. The proposed data augmentation increases tracking accuracy for all but one of the CTC \cite{CTC_old, CTC_23} tested data sets. The $\alpha$ mixing coefficient has been set to the optimal value for each data set. Error bars indicate the standard deviation over three runs.}
\label{tab:tra_results}
\end{table*}

\begin{table}
\centering
\begin{tabular}{@{}l c c c@{}}
\toprule
data set   & $\Delta$ Splits & $\Delta$ FP E & $\Delta$ FN E\\ 
\midrule
Fluo-C2DL-Huh7  & 0 & -1 & -1\\
Fluo-C2DL-MSC   & 0 & -3 & -126 \\
DIC-C2DH-HeLa   & -3 & -2 & -10\\
Fluo-N2DH-GOWT1 & -13 & -3 & -35\\
Fluo-N2DL-HeLa  & -10 & 0 & -17\\
PhC-C2DL-PSC    & -780 & -14 & -230\\
PhC-C2DH-U373   & 0 & 0 & +3\\
\bottomrule
\end{tabular}
%}
\caption{Difference in numbers of tracking mistakes with and without \textit{SynCellFactory}. $\Delta$Split is the difference in splitting detection errors, the $\Delta$FP E represents the difference in false positive edge predictions, and $\Delta$FN E represents the difference in false negative edge predictions. Any value lower than zero signifies an improvement by means of \textit{SynCellFactory}. The most notable improvements from our data augmentation are observed in resolving splitting errors and in reducing missing tracking assignments ($\Delta$FN E). }
\label{tab:tra_results2}
\end{table}

\begin{table}[]
\centering
\begin{tabular}{@{}l c c@{}}
\toprule
data set         & EmbedTrack \cite{kit_track} & Ours \\ 
\midrule
Fluo-C2DL-Huh7  & -                           & \textbf{0.920} \\
Fluo-C2DL-MSC   & 0.693                       & \textbf{0.703} \\
DIC-C2DH-HeLa   & 0.934                       & \textbf{0.943} \\
\bottomrule
\end{tabular}%

\caption{Comparison of the official CTC results. We improved the performance of EmbedTrack \cite{kit_track} by using our data generation for all three data sets. Originally, EmbedTrack \cite{kit_track} was inapplicable to Fluo-C2DL-Huh7 due to a shortage of segmentation masks for training. The generated data sets provided a sufficient amount of segmentation masks, enabling EmbedTrack to track Fluo-C2DL-Huh7.}
\label{tab:ctc_results}
\end{table}

\begin{figure*}
  \centering
      \includegraphics[width=0.99\linewidth]{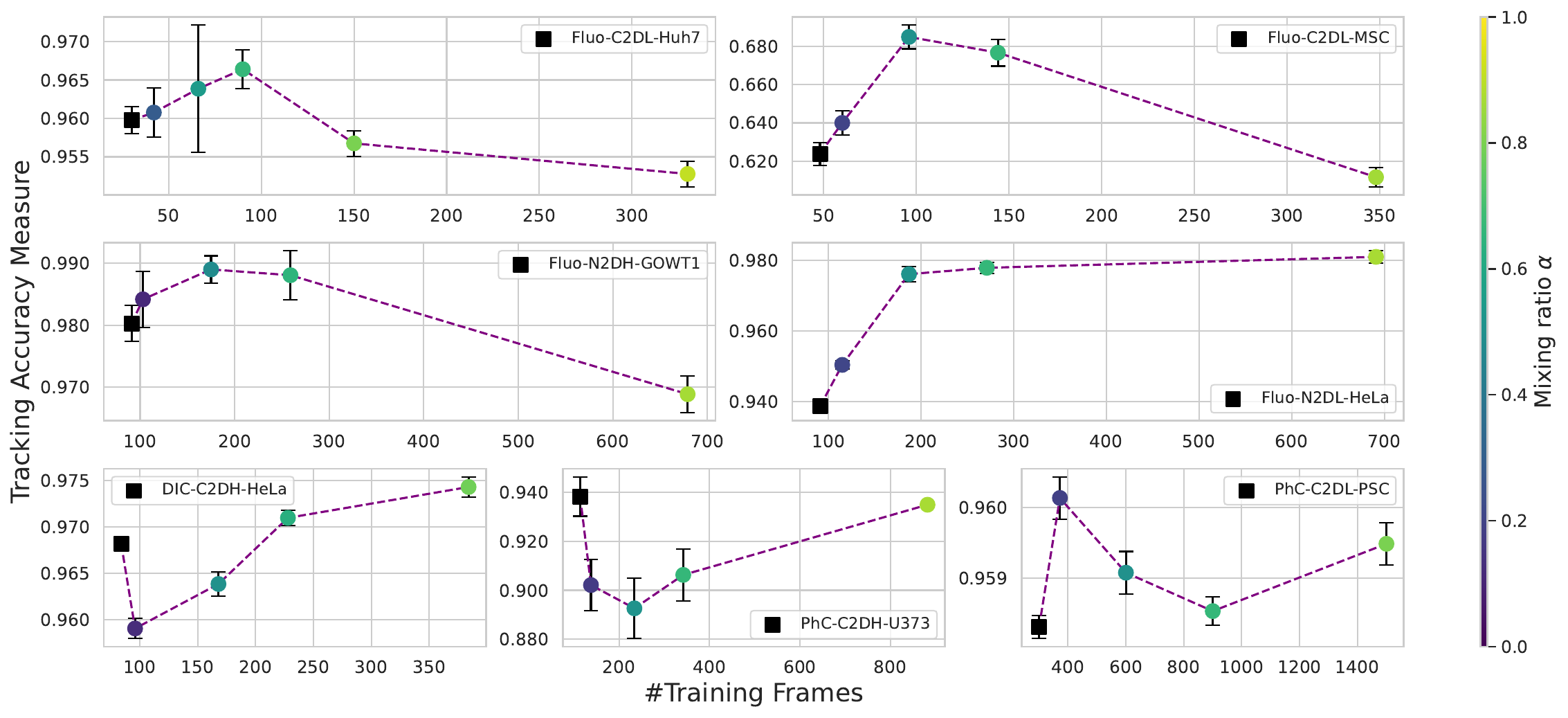}
  \caption{Quantitative results according to the Tracking Accuracy Measure TRA (higher is better). We trained the EmbedTrack model without data augmentation (black square) and with different real and synthetic training data mixing ratios $\alpha$. Here, one can observe that although \textit{SynCellFactory} augmentation positively impacts the TRA score in all but one of the tested data sets, the correct choice of $\alpha$ is critical for the model we benchmarked. Error bars indicate the standard deviation over three runs.}
  \label{fig:results_alpha}
\end{figure*}

\subsection{Quantitative results}
In all but one of the test data sets, our \textit{SynCellFactory} data augmentation improved the tracking quality as measured by the TRA score; the results can be found in \cref{tab:tra_results}.
As shown in previous work using generative models for data augmentation \cite{he2023,eff_diff}, the ratio between synthetic and real data
\begin{equation}
\alpha = \frac{\#\text{syn-frames}}{\#\text{syn-frames} + \#\text{real-frames}}
\end{equation}
in the training set is crucial.
This is also true for the proposed methods; in ~\cref{fig:results_alpha}, we show the tracking accuracy achieved using different mixing ratios $\alpha$.
Our experiments showed two behaviors between fluorescence microscopy data sets (denominated as Fluo-*) and all other microscopy modalities. 
In fluorescence datasets, TRA scores typically increased with the mixing ratio up to an optimal $0.5 < \alpha < 0.7$, then dropped, except in one dataset where improvement continued steadily. In Phase Contrast (PhC-*) and Differential Interference Contrast (DIC-*) experiments, most datasets initially showed a performance drop at low $\alpha$, followed by consistent improvement at high $\alpha \sim 0.8$, with one exhibiting an initial increase, a subsequent drop, and then improvement.
In addition to the TRA score, we also report additional performance metrics in ~\cref{tab:tra_results2} based on the number of correct tracking predictions.

\subsubsection{CTC Results}
In addition to our standard experimental setup, we tested our strategy on the official cell tracking challenge evaluation data set. The CTC organizers evaluate the submissions on a private ground truth. We submitted the results for three data sets. Here, we trained \textit{SynCellFactory} and the EmbedTrack model on all available training data and using the optimal mixing ratio $\alpha$. The results are presented in \cref{tab:ctc_results} and show an improvement on all three data sets compared to those in \cite{kit_track}. 

\subsection{Qualitative Results}
Overall, the results for all tested data sets look realistic from the human perspective. We now present a summary of the qualitative characteristics we observed in our generated images.

\begin{figure}
  \centering
      \includegraphics[width=0.95\linewidth]{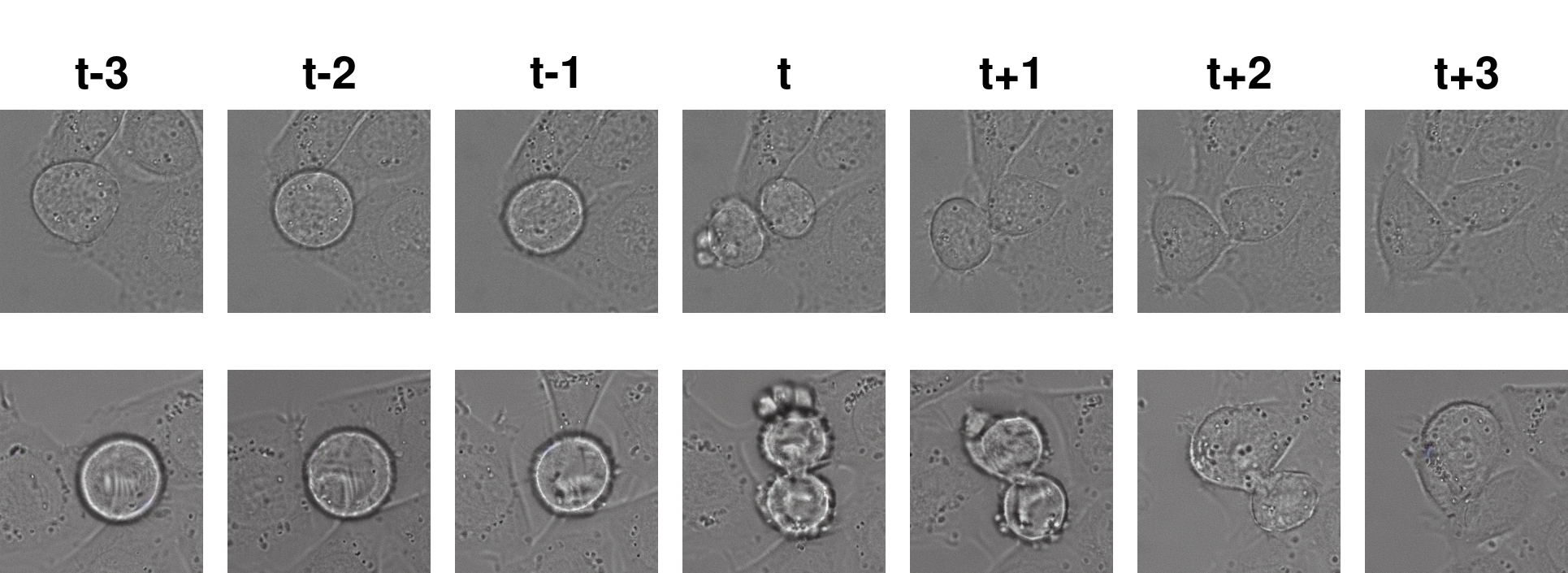}
  \caption{Comparison of real (top) and simulated mitosis (bottom). Shown here are the cell division at time $t$, plus three frames before and after. Key differences include the simulated cell being brighter and lacking the gradual brightness change seen in the real cell before the split. However, the model accurately simulates cell contraction at $t-1$ and realistic artifacts at the split. Post-split, the simulated daughter cells transition realistically from $t+1$ to $t+3$.}
  \label{fig:mito}
\end{figure}

\textbf{Mitosis:} One of the most challenging aspects of automatizing cell tracking is correctly identifying cell splitting. The process by which cells replicate is called mitosis, and most cell types exhibit a drastic change in their appearance during this stage. This process happens in subsequent stages (typically classified into Prophase, Metaphase, Anaphase, and Telophase), each with its distinct visual appearance.
It is, therefore, important to correctly mimic the appearance of cells during mitosis in case visual deep learning models are trained on the generated data sets.
For our data sets, DIC-C2DH-HeLa is the data set with the strongest visual indication of a splitting event. In \cref{fig:mito}, we compare generated and real HeLa cells undergoing mitosis from the DIC-C2DH data set.

\begin{figure}
  \centering
      \includegraphics[width=0.5\linewidth, angle=-90]{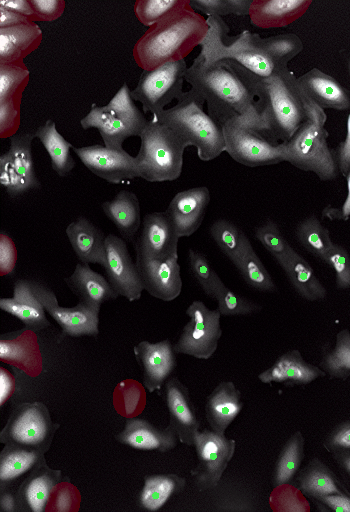}
  \caption{Example of cell hallucinations close to the image boundary. The network inpaints cells marked in red which do not match the conditioning shown in green.}
  \label{fig:hallu}
\end{figure}

\textbf{Hallucinations:} Hallucinations of cells and other artifacts are a weak point of the proposed pipeline (see \cref{fig:hallu} for an example). We observe two main types of hallucinations in the generated data sets.
The first mode of hallucination is the appearance of additional cells near image borders. These cells are usually inpainted from the beginning of the video and remain stationary throughout the video. 
We hypothesize that these hallucinations are derived from the lack of cell detections at the border in the CTC data sets. Indeed, cells partially cut out of the field of view are often not included in the ground truth.
The second type of hallucination occurs more sporadically and consists of artifacts and debris in the generated frames seldomly promoted to cells by our ControlNet. We observed such hallucinations more prevalently in data sets with much background noise, like Fluo-C2DL-MSC and PhC-C2DH-U373. 

%% --- Conclusion ---
\section{Conclusion}
Despite the positive results, \textit{SynCellFactory} is not without limitations, which also serve as directions for future work.

The current motion module in \textit{SynCellFactory} is simplistic, focusing on basic cell movements. This limits its applicability in capturing the full spectrum of biological diversity and complexity. Future iterations of \textit{SynCellFactory} could benefit from a more sophisticated motion module that can accurately model a broader range of biological behaviors and interactions.
Moreover, the model's performance declines in scenarios with high cell density or dramatic changes in cell count. Addressing this limitation would enhance the model's utility in more complex biological environments.
While \textit{SynCellFactory} does not produce segmentation ground truth, the availability of advanced segmentation tools partially mitigates this issue. However, it remains a potential source of inaccuracies and could be an area for future integration.
\textit{SynCellFactory} can sample videos of arbitrary length, but there is a noticeable drop in quality for extended sequences; in our experiments, we found that after 30 frames, the quality is reduced to the point of losing most resemblance with the original data. We hypothesize that this is due to the training procedure, in which the CN-Mov is only trained on frames pair and real data conditioning. Future developments could focus on maintaining consistent quality across varying video lengths.

In conclusion, \textit{SynCellFactory} represents a significant step forward in the field of biological data augmentation. Its ability to generate realistic and diverse data sets holds great potential for advancing deep learning-based cell tracking pipelines. While there are several areas for improvement, the foundation laid by \textit{SynCellFactory} offers a solid base for the concrete application of generative AI in biomedical computer vision.

\section{Code Availability}
The code supporting this work is published and freely available at the following URL: \href{https://github.com/mosturm/SynCellFactory}{https://github.com/mosturm/SynCellFactory}.

{
    \small
    \bibliographystyle{ieeenat_fullname}
    \bibliography{main}
}

\clearpage
\setcounter{page}{1}
\maketitlesupplementary

\appendix
\section{Motion Model}
\subsection{Statistics}
In our study, we derive the motion module's key statistics from both the tracking ground truth and a limited set of segmentation ground truth data.

We use all segmentation masks available within a single data set to calculate the mean $A_c$ and standard deviation $\sigma_A$ of pixel counts per segmentation mask.

We estimate our cell displacement distribution given the tracking ground truth by computing the displacements of individual cells between consecutive timeframes. The distribution of all calculated displacements is then fitted using a gamma distribution $G$ with respect to its parameters $\alpha_d, \Theta_d$:
\begin{equation} G(x; \alpha_d, \Theta_d) = \frac{x^{\alpha_d-1} \cdot e^{-\frac{x}{\Theta_d}}}{\Theta_d^{\alpha_d} \cdot \Gamma(\alpha_d)} \end{equation}
Here, $\alpha_d$ represents the shape parameter, $\Theta_d$ the scale parameter, and $\Gamma(\alpha_d)$ denotes the gamma function. The choice of a gamma distribution is empirically driven by its adaptability in modeling data with skewed distributions.

We can generate new, plausible samples of cell population timelapses with these statistical parameters determined. These samples take into account variations in both cell area and motion characteristics.

\subsection{Simulation}
\noindent \textbf{Initialisation}: 
In the first timeframe we sample $n$ cell centers at random positions $(x_i,y_i) \in (0,1) $ for $i \in \{1, 2, \ldots, n\}$ with each cell having an area $A_i \sim \mathcal{N}(A_c,\frac{\sigma_A}{10})$. We model each cell as a 2D disk with radius $r_i=\sqrt{\frac{A_i}{\pi}}$.

\noindent \textbf{Movement of cells}:
In our simulation, cell movements are modeled as a random walk. 

Let \(\mathbf{p}_{i, t} = (x_{i, t}, y_{i, t})\) represent the position vector of cell \(i\) at timeframe \(t\). The position at the next timeframe \(t + 1\) is updated using:
\begin{align}
\phi &\sim \text{Uniform}(0,2\pi) \\
m_v &\sim G(\alpha_d, \Theta_d) \\
\mathbf{p}_{i, t+1} &= \mathbf{p}_{i, t} + m_v 
\begin{pmatrix}
\cos(\phi) \\
\sin(\phi)
\end{pmatrix}  
\end{align}

Each cell has a splitting probability \(p_{\text{split}}\) at each timeframe. Upon splitting, a cell divides into two daughter cells along a random axis. The position vectors of the daughter cells, denoted as \(\mathbf{p}_{i,1}\) and \(\mathbf{p}_{i,2}\), are given by:
\begin{equation}
  \mathbf{p}_{i,1/2} = \mathbf{p}_i \pm d \cdot r_i 
\begin{pmatrix}
\cos(\phi) \\
\sin(\phi)
\end{pmatrix}  
\end{equation}

Here, \(d \sim \text{Uniform}(0.7,0.8)\) introduces a slight randomness in the splitting process, and \(r_i\) is a scaling factor.

\noindent \textbf{Resolving overlaps}:
The resolution of overlaps in our model is repeated in successive iterations, until no overlaps remain.

To resolve the overlap of two circles, we first define the repulsion vector $\mathbf{v}_{\text{repulsion}}$ between two overlapping circles $c_i$ and $c_j$ with center coordinates $(x_i,y_i)$ and $(x_j,y_j)$ as follows:
\begin{equation}
 \mathbf{v}_{\text{repulsion}} = \begin{pmatrix} x_i - x_j \\ y_i - y_j \end{pmatrix} 
 \label{eqn:vrep}
\end{equation}

The angle of repulsion $\theta$ is calculated using the arctan function:
\begin{equation}
 \theta = \arctan((\mathbf{v}_{\text{repulsion}})_y, (\mathbf{v}_{\text{repulsion}})_x) + \theta_{\epsilon},
\end{equation}
where $\theta_{\epsilon} \sim \text{Uniform}(-0.1, 0.1)$ is a small perturbation to the angle to prevent infinite loops during the overlap resolution process.     

Next, we calculate the repulsion displacement $F_{i,j}$, based on the overlap distance between the circles, scaled by the average of their radii to account for potentially larger overlap areas in bigger circles. The formula is given as:
\begin{equation}
        F_{i,j} = \left(\left( r_i + r_j - \|\mathbf{v}_{\text{repulsion}}\| \right) \cdot \frac{r_i + r_j}{2} \right) \cdot \left(1 + \epsilon \right)
        \label{eqn:Frep}
\end{equation}
where $\epsilon \sim \text{Uniform}(0,0.2)$ is introduced in each iteration with the same reasoning as for $\theta_{\epsilon}$.

The positions of the cells are then updated as follows:
\begin{align}
    \mathbf{p}_{i} &= \mathbf{p}_{i} + F_{i,j} \cdot 
    \begin{pmatrix}
    \cos(\theta) \\
    \sin(\theta)
    \end{pmatrix} \\
    \mathbf{p}_{j} &= \mathbf{p}_{j} - F_{i,j}\cdot 
    \begin{pmatrix}
    \cos(\theta) \\
    \sin(\theta)
    \end{pmatrix}
    \label{eqn:resolve}
\end{align}

\newpage

\section{ControlNet Conditioning}
\begin{figure}[!h]
  \centering
      \includegraphics[width=0.65\linewidth]{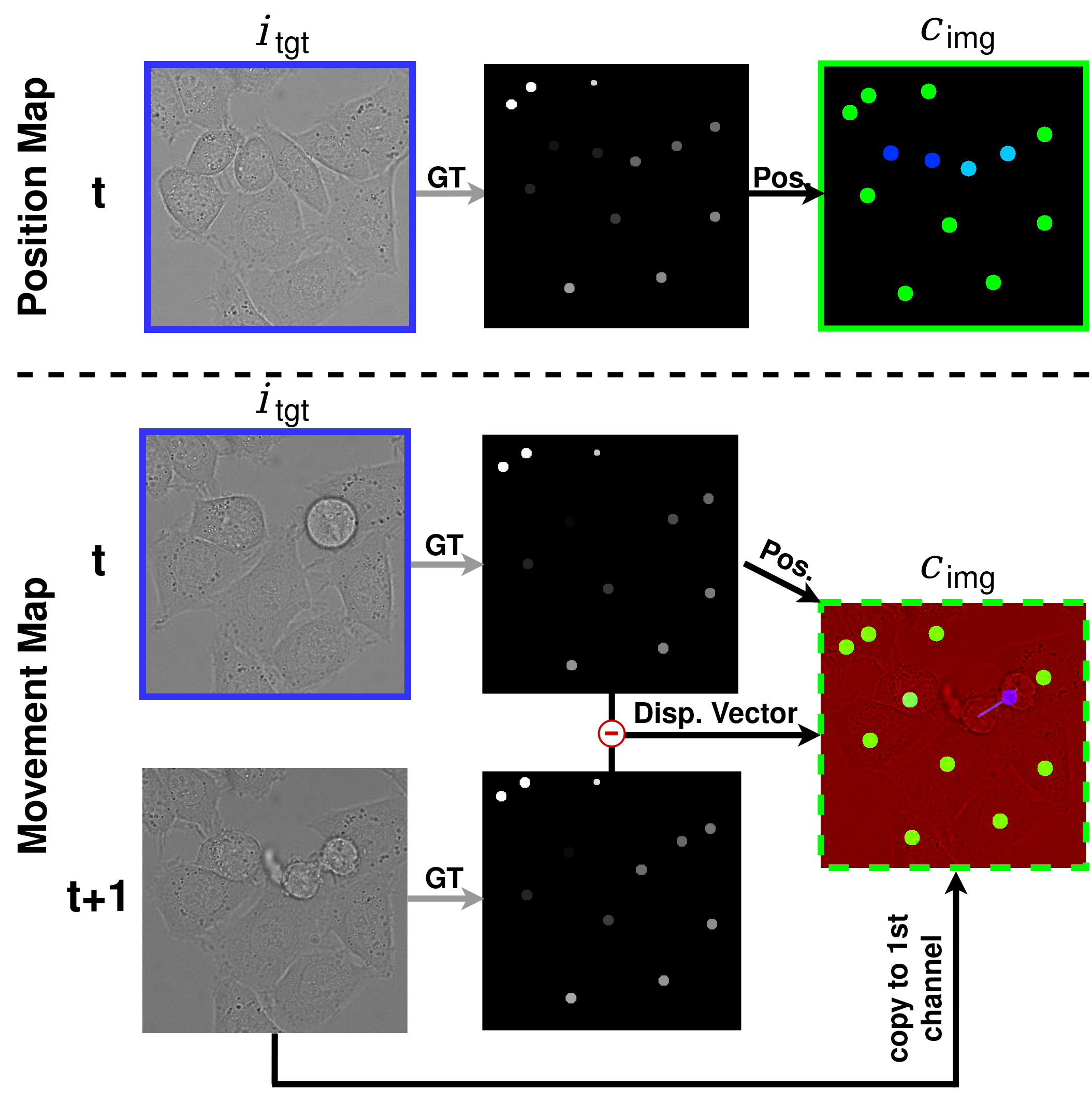}
  \caption{Construction of training pairs [$c_{\text{img}}$ (green squared images), $i_{\text{tgt}}$ (blue squared images)] from real cell videos. The network is then trained with [$c_{\text{img}}$, $i_{\text{tgt}}$] as depicted in Fig. \ref{fig:cnet_detailed}.\\
  \textit{Top:} Construction of a position map for training. Given a real timeframe, we know the cell positions from the corresponding detection ground truth (GT). We then transform the detection ground truth into our color-coded position map as described in \cref{subsec:cn-pos}. \\
  \textit{Bottom:} Construction of a movement map for training. The goal is that the network learns to produce a realistic-looking timeframe $t$, given the appearance of timeframe $t+1$ and individual cell displacements. We construct the movement map in the following way:
  We overlay timeframe $t+1$ with a position map from timeframe $t$ and displacement vectors computed from the tracking ground truth. With this setup, the network can learn to draw a realistic timeframe $t$ by learning where to draw cells (position map of $t$), where they were previously located in $t+1$ (displacement vectors), and how they looked like in $t+1$. }
  \label{fig:cons}
\end{figure}

In \cref{fig:cons} we showcase an example of the construction of position and movement maps for training. 

\vspace*{-0.58cm}
\section{Segmentation Pseudo Ground Truth}
\begin{figure}[!h]
  \centering
      \includegraphics[width=0.95\linewidth]{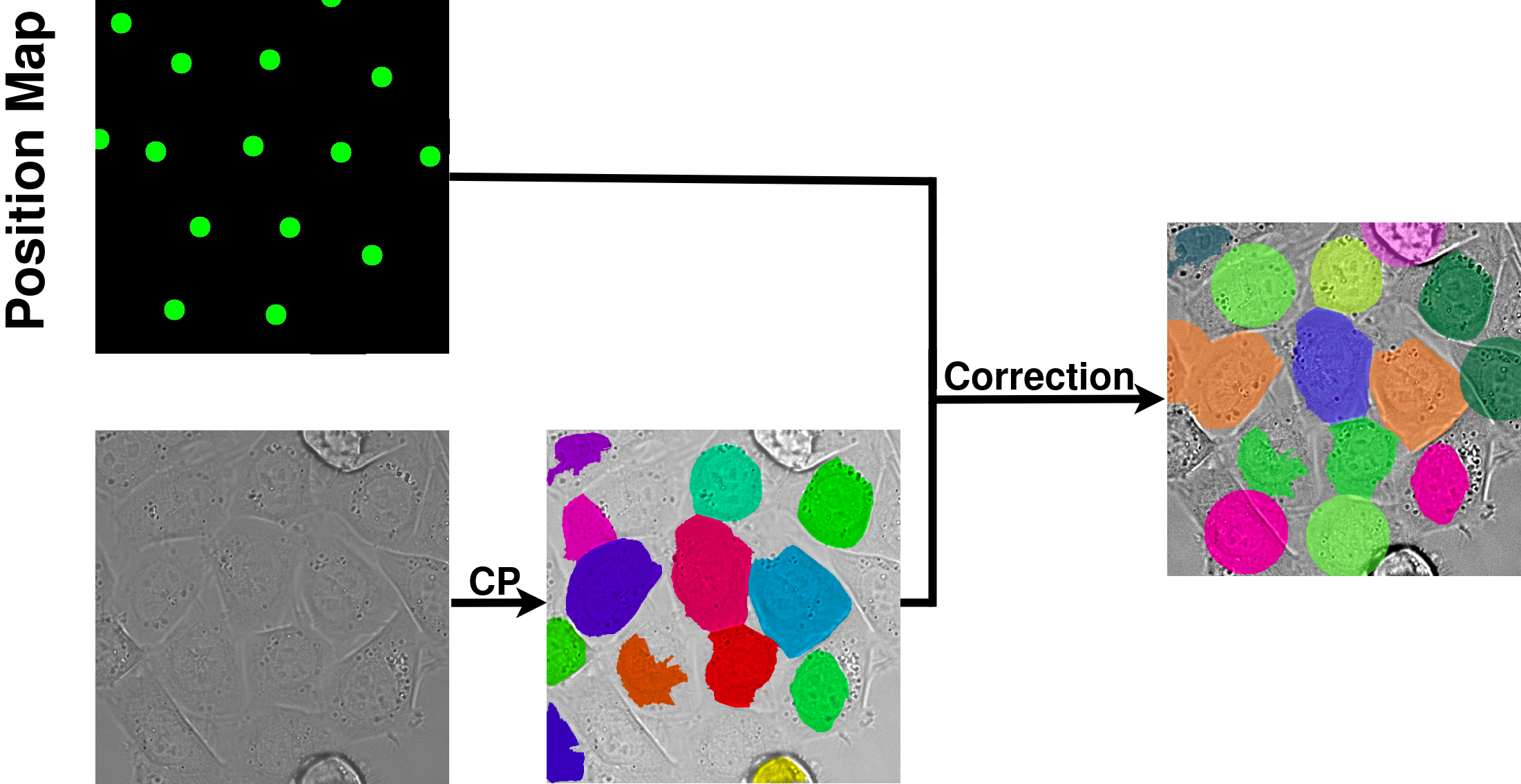}
  \caption{Calculation of approximate segmentation masks for our generated time-frames.
  We use finetuned Cellpose (CP) \citep{cp1,cp2} segmentation models to predict segmentation masks. These masks are corrected using the corresponding position map; for more details, see \cref{subsec:seggt}.  }
  \label{fig:corr}
\end{figure}

The segmentation correction step is illustrated in \cref{fig:corr}.
\noindent
\section{ControlNet Training}
To address the challenges of limited training data we found it beneficial to choose a sequential training procedure, training in total four consecutive ControlNets per dataset.

We start our training procedure by training a base model (BM) of CN-pos on randomly selected patches of $c_{\text{img}}$ and $i_{\text{tgt}}$ measuring $h/2 \times w/2$. The initial phase focuses on enabling ControlNet to learn the correlation between position maps and cell appearance.

After a fixed number of iteration steps (see \cref{tab:train_para}) the training procedure of CN-pos is stopped and we start training an initial version of CN-mov. This phase involves using cropped movement maps and corresponding target images of size $h/2 \times w/2$, similarly to the base model of CN-pos.

CN-mov is initialized with the pre-trained weights from the CN-pos base model, ensuring continuity in the learning process.

The final step in our training methodology involves fine-tuning both base model ControlNets. This is achieved through limited optimization steps conducted on the full-sized $h \times w$ training timeframes. After fine-tuning, the full model (FM) ControlNets are optimized and ready to integrate into the sampling process.

In \cref{tab:train_para} is a list of training parameters used to train our ControlNets. Numbers for one A100 40 GB GPU.

\begin{table}[h!]
\centering
\small  % Reducing the font size
\begin{tabular}{@{}l c@{}}
\toprule
Total Trainable Parameters                       & 1.2 B                                        \\ 
Latent Representation                           & shape(z) = (1,4,64,64)                       \\
Diffusion Steps                                 & 50                                           \\
Optimizer                                       & AdamW                                        \\
Learning Rate                                   & 5e-6                                         \\
Batch Size                                      & 4                                            \\
Training BM CN-pos  $n_{\text{cell}}: 10^2; 10^3$    & 30000; 60000 Steps                         \\
Training BM CN-mov  $n_{\text{cell}}: 10^2; 10^3$    & 10000; 20000 Steps                         \\
Finetuning FM CN-pos $n_{\text{cell}}: 10^2; 10^3$  & 3000; 7000 Steps                           \\
Finetuning FM CN-mov  $n_{\text{cell}}: 10^2; 10^3$ & 3000; 7000 Steps                          \\
Sampling Time (12 Timeframes)                   & $\sim$ 3 min                                 \\
\bottomrule
\end{tabular}
\caption{Parameters used for ControlNet training and sampling. 
$n_{\text{cell}}$ denotes the number of cells expected in a single timeframe and determines the number of optimization steps. For example, if $n_{\text{cell}}$ is in the order of $10^1 - 10^2$, we start our training procedure by training the CN-pos base model (BM) for 30000 optimizer steps. If $n_{\text{cell}}$ is in the order of $10^2 - 10^3$, we train the CN-pos base model for 60000 optimizer steps. The optimizer steps for consecutive ControlNet training are denoted in the same way in this table.   }
\label{tab:train_para}
\end{table}

\vspace{-0.58cm}

\section{Additional Generated Time-Lapse}
\begin{figure*}
  \centering
      \includegraphics[width=0.99\linewidth]{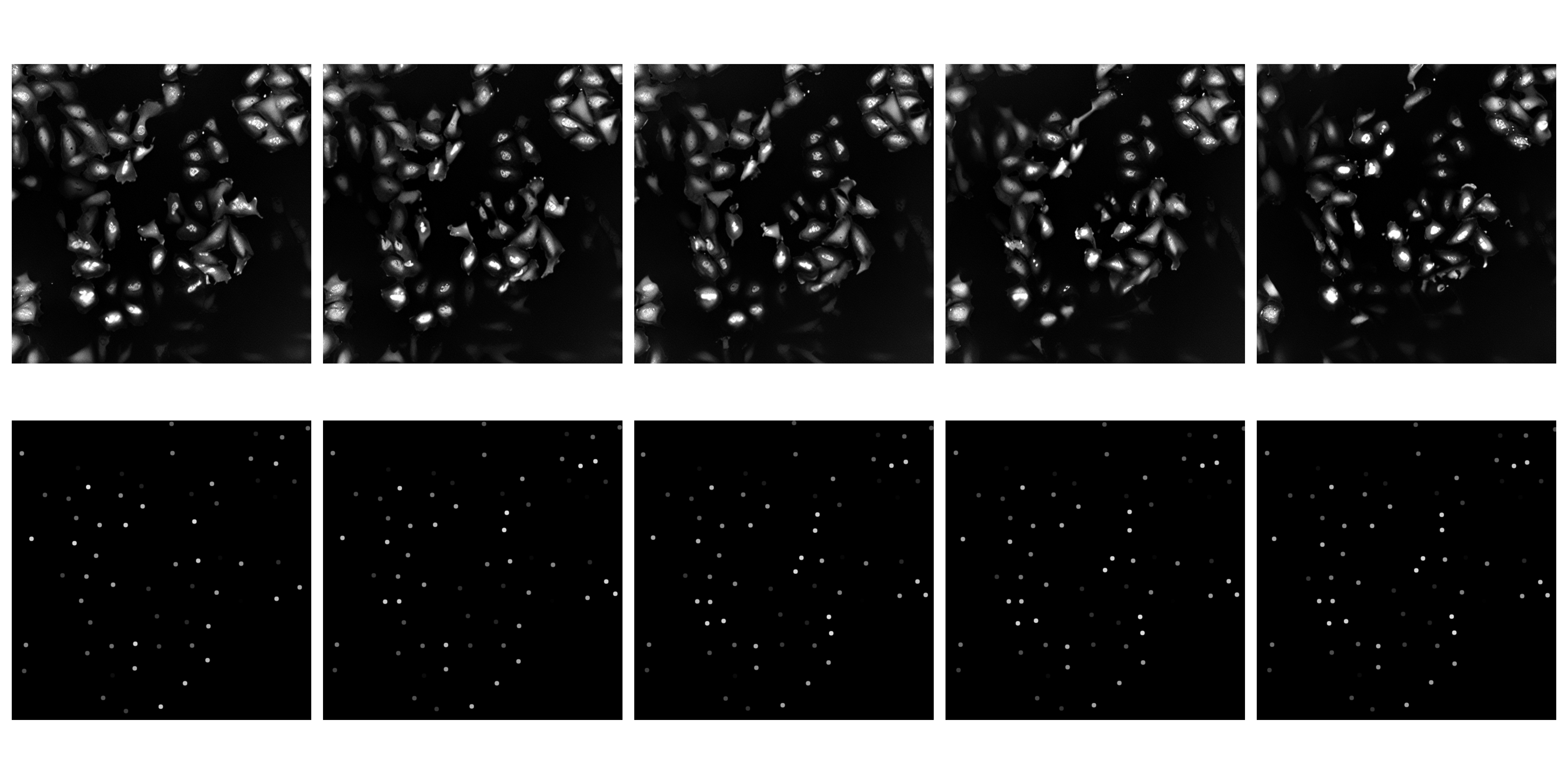}
  \caption{Fluo-C2DL-Huh7}
  \label{fig:HUH_supp}
\end{figure*}

\begin{figure*}
  \centering
      \includegraphics[width=0.99\linewidth]{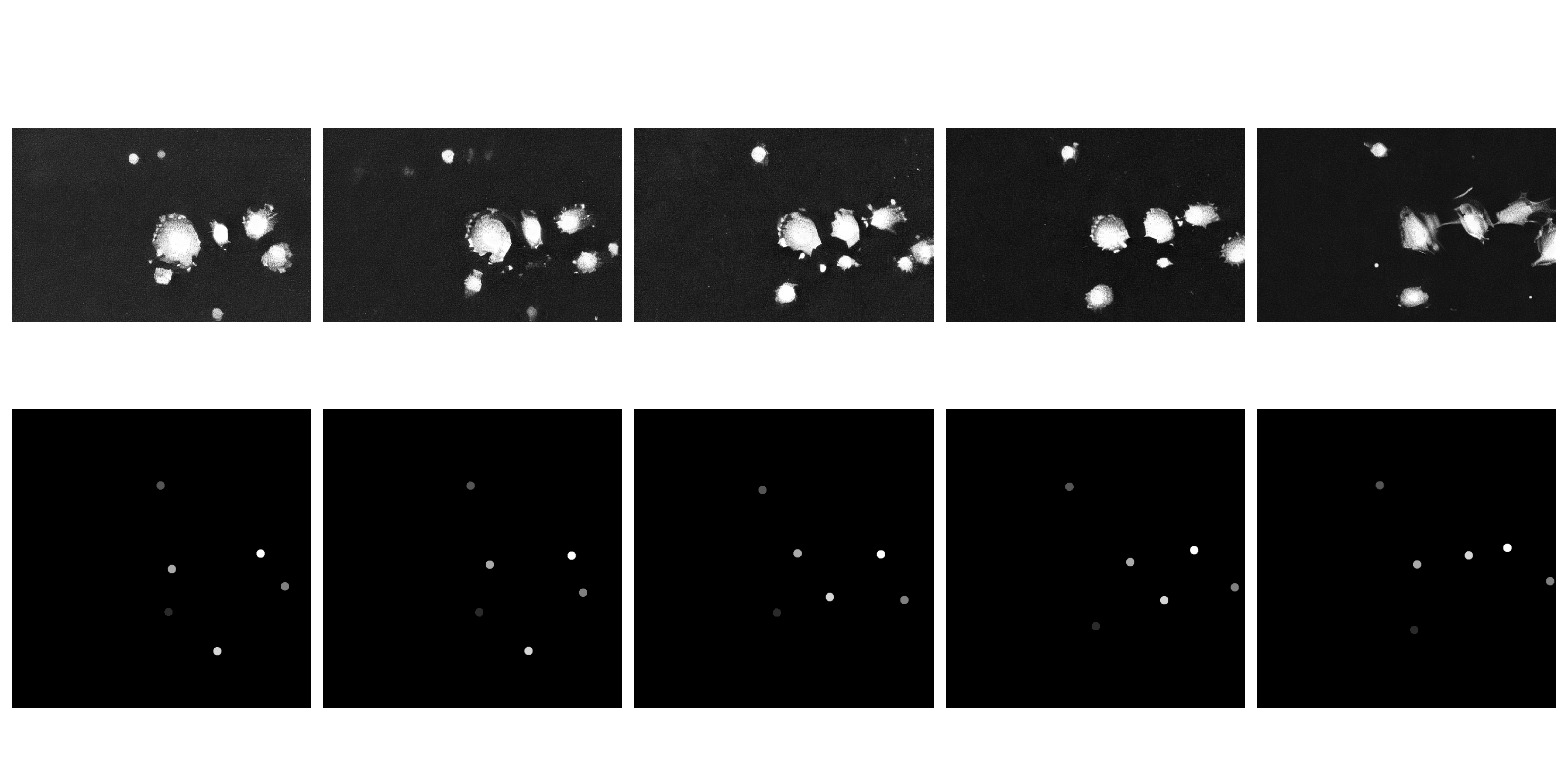}
  \caption{Fluo-C2DL-MSC}
  \label{fig:MSC_supp}
\end{figure*}

\begin{figure*}
  \centering
      \includegraphics[width=0.99\linewidth]{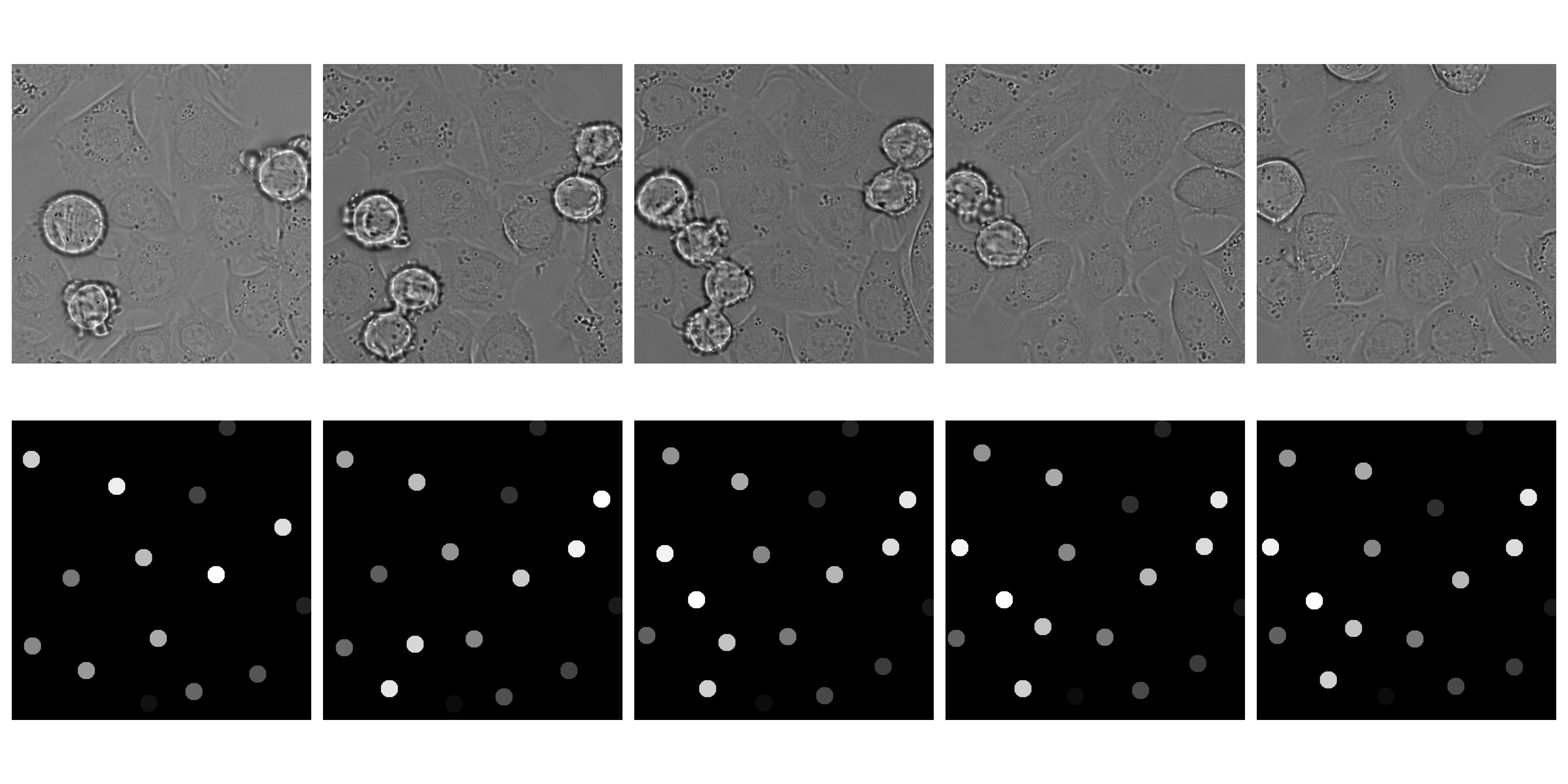}
  \caption{DIC-C2DH-HeLa}
  \label{fig:HELA_supp}
\end{figure*}

\begin{figure*}
  \centering
      \includegraphics[width=0.99\linewidth]{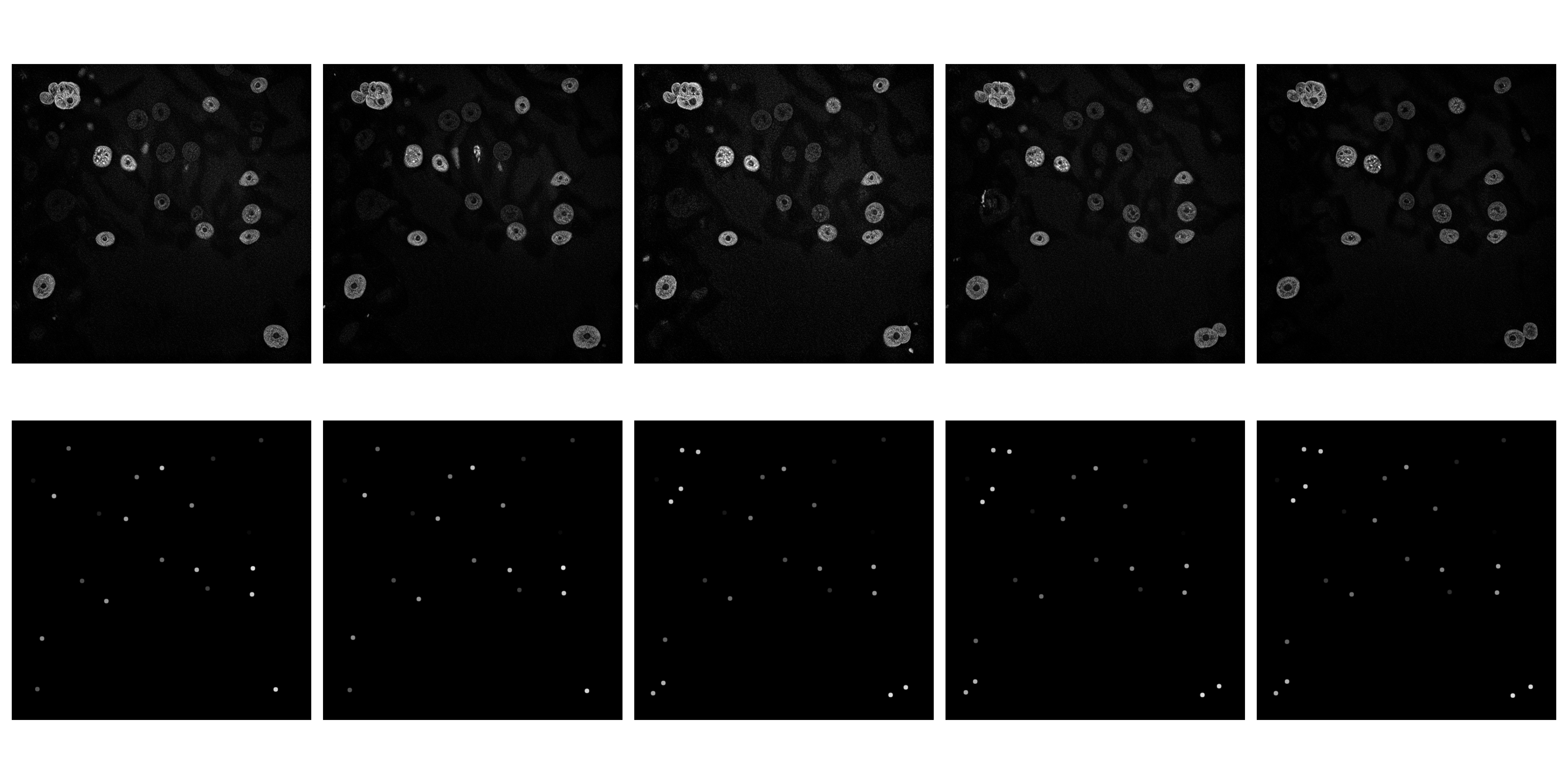}
  \caption{Fluo-N2DH-GOWT1}
  \label{fig:GOWT_supp}
\end{figure*}

\begin{figure*}
  \centering
      \includegraphics[width=0.99\linewidth]{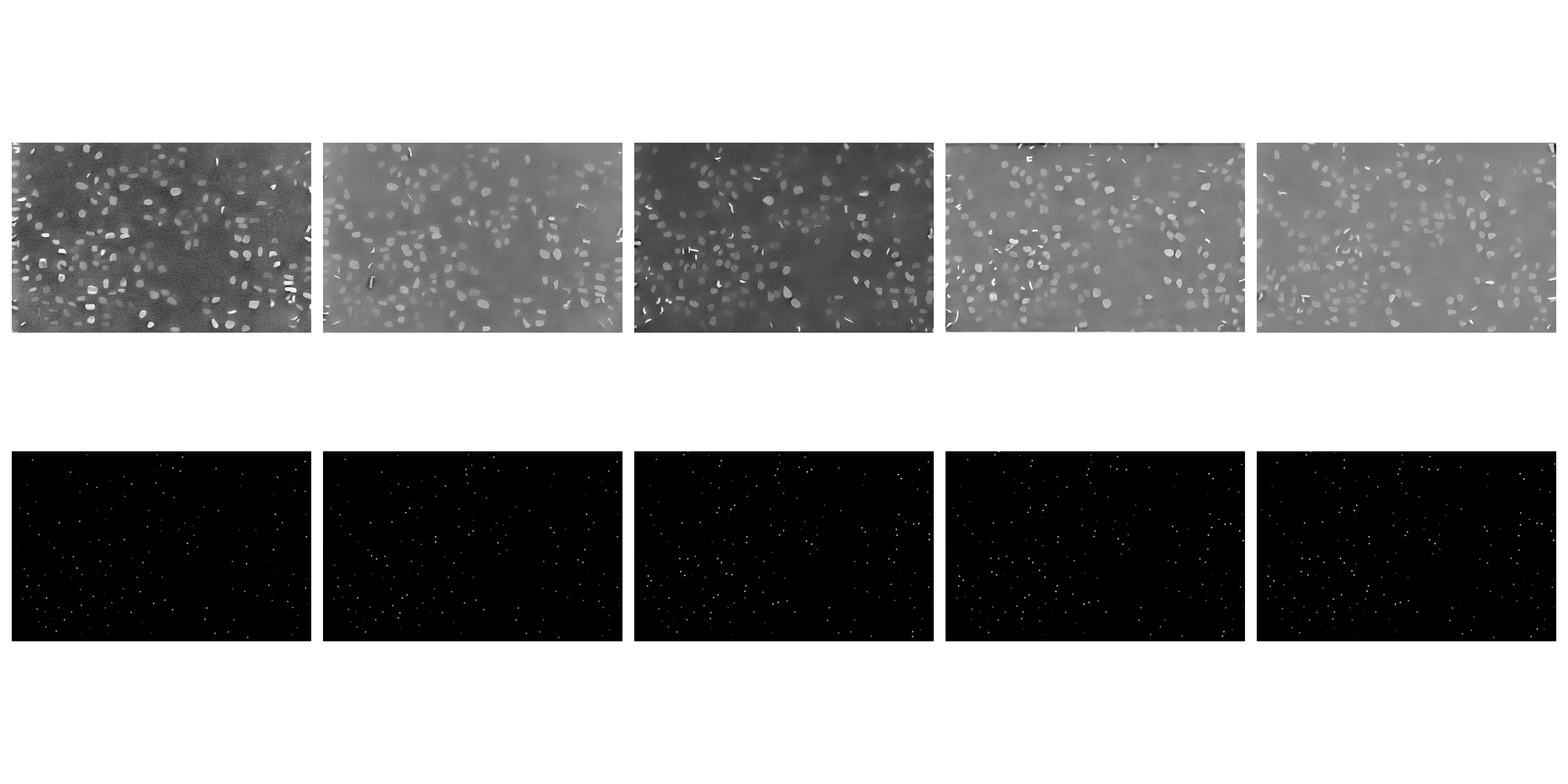}
  \caption{Fluo-N2DL-HeLa}
  \label{fig:N2DL_supp}
\end{figure*}

\begin{figure*}
  \centering
      \includegraphics[width=0.99\linewidth]{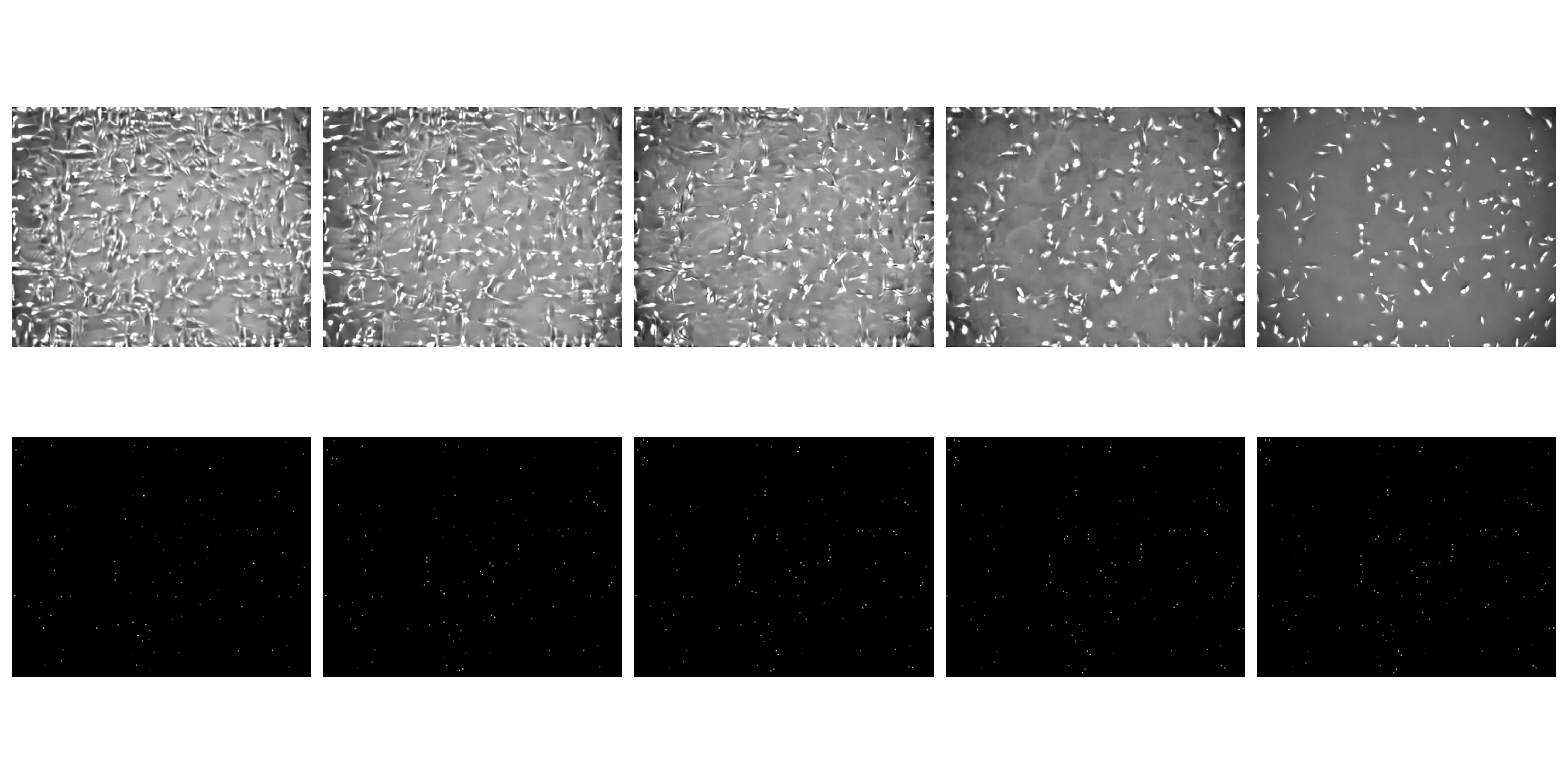}
  \caption{PhC-C2DL-PSC}
  \label{fig:PSC_supp}
\end{figure*}

\begin{figure*}
  \centering
      \includegraphics[width=0.99\linewidth]{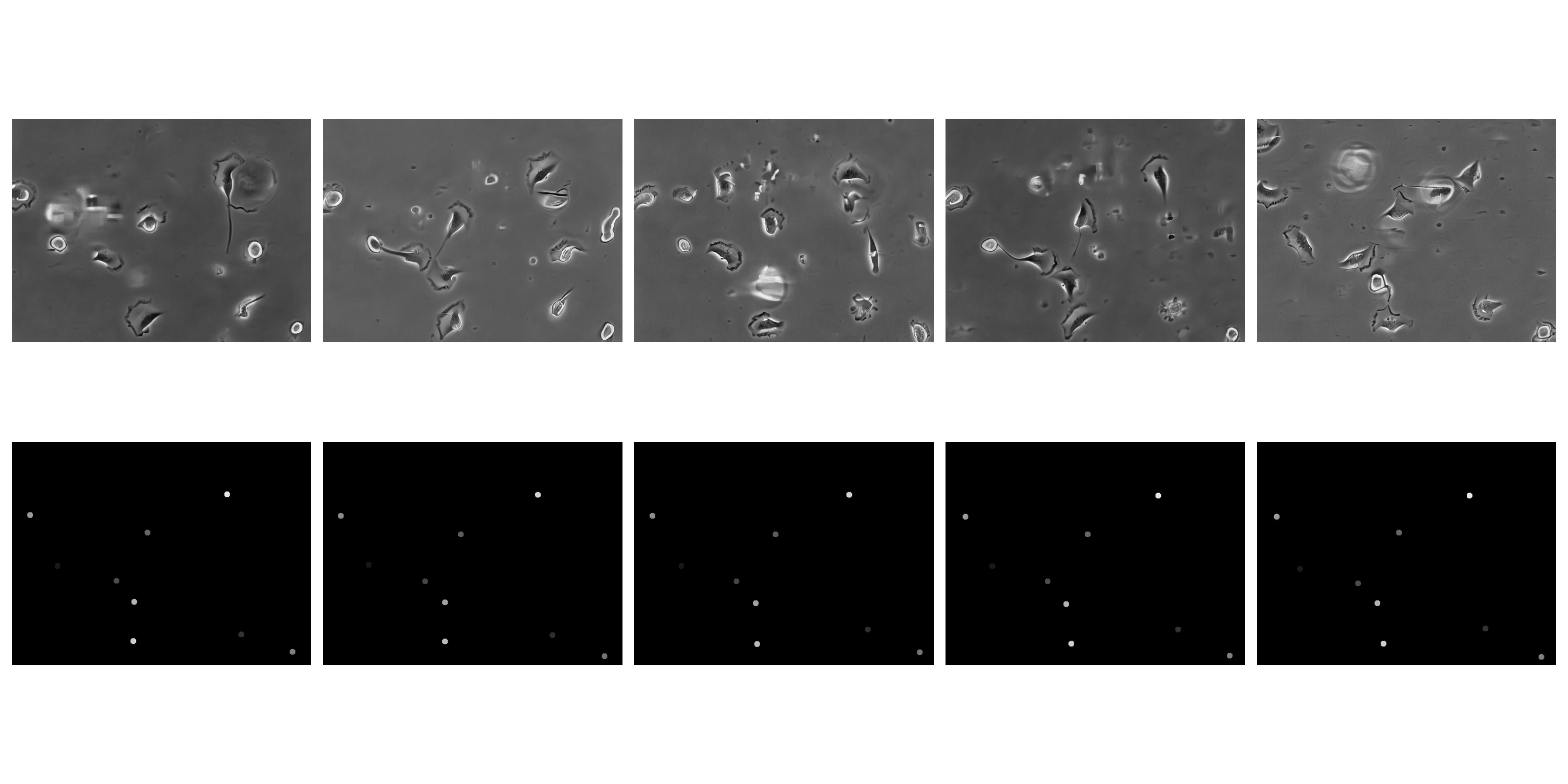}
  \caption{PhC-C2DH-U373}
  \label{fig:U373_supp}
\end{figure*}

In \cref{fig:HUH_supp} - \cref{fig:U373_supp}, we showcase more examples of generated time-lapses and the corresponding tracking ground truth as simulated by the motion module. Generated time-frames are shown in the top row, and tracking ground truth is shown in the bottom row. 

\end{document}